\date{}
\documentclass{article}

\usepackage[linesnumbered,ruled]{algorithm2e}
\usepackage{amsmath} 
\usepackage[a4paper]{geometry}
\usepackage{apacite}
\usepackage{natbib}
\usepackage{float}
\usepackage{graphicx} 
\usepackage{latexsym,amsmath,amsfonts,amsthm,amssymb,graphicx,color,mathrsfs}
\usepackage{xcolor}
\usepackage{subfigure}
\usepackage{subcaption}
\usepackage{tabularx}
\usepackage{booktabs}
\usepackage{bm}
\usepackage{tikz}
\usepackage{multirow} 
\usepackage{adjustbox}
\usepackage{array} 
\usepackage{listings}%
\usepackage{amssymb}
\numberwithin{equation}{section}
\theoremstyle{plain}
\newtheorem{thm}{\protect\theoremname}[section]
\theoremstyle{plain}

\theoremstyle{definition}

\usepackage{caption}
\usepackage[nottoc]{tocbibind}
\allowdisplaybreaks[4]
\usepackage{dsfont}
\DeclareMathAlphabet{\mathcal}{OMS}{cmsy}{m}{n}

\providecommand{\lemmaname}{Lemma}

\providecommand{\theoremname}{Theorem}

\makeatother

\providecommand{\definitionname}{Definition}
\providecommand{\lemmaname}{Lemma}
\providecommand{\theoremname}{Theorem}

\usepackage{authblk}

\begin{document}
\title{Rough Path Signature-Guided Geometry Augmentation for Few-Shot Industrial Surface Defect Detection}

\author{Jiaqi Kuang\\
Oxford Suzhou Centre for Advanced Research, University of Oxford\\
Suzhou, China\\
\texttt{jiaqi.kuang@oscar.ox.ac.uk}}

\maketitle
\begin{abstract}
Few-shot industrial defect detection remains difficult for standard supervised detectors, which achieve poor performance on boundary-dominated industrial defects.

This paper proposes rough path signature-guided geometry augmentation (RPS-GA), a geometry-aware approach in which Canny edge contours are treated as ordered planar paths whose truncated second-order signature responses, especially the antisymmetric L\'evy-area term, are aggregated into a spatial map that highlights boundary-related structure through two fusion operators, SIG-AUG and SGAA.
The approach is evaluated on NEU-DET and PCB-Defect under a few-shot protocol with 5, 10, 20, or 50 labeled images per class, using an unmodified YOLOv8n detector throughout.
Compared with the baseline, RPS-GA delivers large gains when supervision is limited, although the margin shrinks as more labels become available.
On NEU-DET, SIG-AUG raises 10-shot mAP@0.5 from 0.341 to 0.583, whereas on PCB-Defect, SGAA improves 10-shot mAP@0.5 from 0.086 to 0.299 and yields usable detection at 5-shot where the baseline fails entirely.
These trends are confirmed by multi-seed evaluation across independent random partitions.
Overall, the results indicate that second-order path-signature geometry offers a practical way to strengthen few-shot industrial defect detection without meta-learning or detector redesign.
\end{abstract}

\medskip{}

\newpage

\section{Introduction}
\label{sec:introduction}

Automated visual inspection is a core component of quality control in manufacturing.
In metal finishing, electronic assembly, and related production lines, surface and PCB defects must be detected early to limit scrap, rework, and downstream reliability failures~\citep{tabernik2020segmentation,huang2019pcb}.
Deep learning detectors, especially one-stage architectures such as YOLO~\citep{redmon2016you,sapkota2025ultralytics}, have therefore become widely adopted because they offer a favorable balance between accuracy and inference speed on industrial hardware.

Most reported inspection systems nevertheless assume access to hundreds or thousands of labeled images per defect class.
In practice, inspection must often be deployed after only a short pilot run: new products, rare failure modes, and line changeovers routinely leave engineers with only a handful of annotated examples per category.
Under such a few-shot scenario, standard supervised training on a fixed detector often fails to detect thin, low-contrast, or small defects reliably, although the same architecture achieves good performance when sufficient labeled data are available~\citep{he2019end,huang2019pcb}.
Improving performance in this regime without redesigning the deployed inspection system remains an important practical challenge.

Existing literature follows two largely separate directions.
First, few-shot and meta-learning methods address label scarcity by changing how the detector is trained or structured.
Representative strategies include episodic training, metric learning, and modified detection architectures~\citep{kang2019few,wu2020meta,yan2019meta,fan2020few}.
Although effective on some natural-image benchmarks, these approaches often require carefully designed base/novel splits or non-standard training pipelines that are difficult to integrate with established YOLO-based workflows on the factory floor.

Second, a complementary line of work keeps the detector fixed and instead enriches the \emph{training data} through augmentation~\citep{shorten2019survey,ghiasi2021simple}.
Random photometric and geometric transforms, together with copy-paste-style instance mixing, increase data diversity and are easy to integrate into standard detection pipelines~\citep{ghiasi2021simple,ge2021yolox}.

However, generic augmentations do not explicitly encode the boundary geometry of defects.

Industrial defects are often defined by elongated contours, local curvature, and boundary continuity on repetitive backgrounds, rather than by distinctive object-level appearance alone.
Existing edge-based or attention-based enhancements therefore remain largely filter-driven and do not provide a path-level description of boundary shape~\citep{tabernik2020segmentation}.

Neither direction fully resolves few-shot industrial inspection under a fixed deployment setting.
Rough path signatures offer a principled geometric summary of trajectories~\citep{lyons1998,lyons2007differential}, but they have rarely been used to generate label-preserving training views for image-based few-shot defect detection.

Therefore, this paper proposes \emph{rough path signature-guided geometry augmentation} (RPS-GA), a data-centric framework for few-shot industrial surface defect detection.
In RPS-GA, ordered edge contours are first extracted from each training image with Canny detection and then treated as planar paths for which a truncated second-order path signature is computed.
The antisymmetric L\'evy-area response of each retained contour is aggregated into a single spatial map $M(\mathbf{I})$ that covers the full image domain.
Two label-preserving fusion operators, SIG-AUG and SGAA, build an additional training view from $M(\mathbf{I})$ without changing the bounding-box labels, where SIG-AUG applies a pseudo-color overlay and SGAA applies multiplicative attention to emphasize boundary-related geometry.
All experiments are conducted with an unmodified YOLOv8n detector, which allows RPS-GA to be evaluated as a data-construction module that is compatible with existing YOLO-based inspection workflows.

Experiments are conducted on two public benchmarks---NEU-DET for hot-rolled steel surface defects and PCB-Defect for PCB assembly defects---under a few-shot protocol with $K\in\{5,10,20,50\}$ labeled images per class.
For each selected training image, RPS-GA adds one signature-guided view with unchanged bounding-box labels; YOLOv8n is trained without architectural modification, and all reported test results are obtained on original images only.

Results show that signature-guided augmentation is most beneficial when labels are scarce.
On NEU-DET, both SIG-AUG and SGAA clearly outperform the YOLOv8n baseline across 5-, 10-, and 20-shot settings; SIG-AUG reaches 0.583 mAP@0.5 at 10-shot compared with 0.341 for the baseline, and similar gains are observed under three independent random partitions.
On PCB-Defect, the baseline essentially fails at 5-shot (mAP@0.5 = 0.000), whereas RPS-GA yields usable detection; at 10-shot, SGAA improves mAP@0.5 from 0.086 to 0.299.

Our study contributes to the existing literature in the following respects:
First, RPS-GA achieves substantial few-shot detection gains on a fixed YOLOv8n detector without meta-learning or architectural redesign, showing that severe label scarcity in industrial inspection can be mitigated within a standard supervised detection workflow.
Second, it develops a geometry-aware augmentation paradigm for defect images by encoding boundary structure through contour-based, second-order path-signature responses, thereby refining geometry-oriented few-shot learning beyond generic photometric transforms and filter-based edge enhancement.
Third, it introduces rough path signature theory into few-shot industrial surface defect detection and demonstrates a new image-based application of signature methods, extending rough-path geometric analysis to visual inspection.

The structure of this paper is organized as follows:
Section~\ref{sec:literature_review} reviews related work on industrial defect detection, few-shot object detection, data augmentation, and path signatures.
Section~\ref{sec:theory} summarizes the truncated signature and second-order L\'evy-area decomposition used to motivate the response map.
Section~\ref{sec:method} presents the RPS-GA pipeline, datasets, and few-shot training protocol.
Section~\ref{sec:results} reports quantitative and qualitative results on NEU-DET and PCB-Defect.
Section~\ref{sec:conclusion} concludes the paper and discusses limitations.

\section{Literature Review}
\label{sec:literature_review}

Few-shot industrial defect detection sits at the intersection of visual inspection, low-data learning, and data augmentation.
Existing work progresses along four largely separate lines: deep learning for surface and PCB inspection, few-shot object detection, augmentation strategies for training data-scarce detectors, and path-signature methods for sequential and geometric data.
This section reviews each line and identifies the gap addressed by rough path signature-guided geometry augmentation (RPS-GA).

\subsection{Deep learning for industrial surface defect detection}

Convolutional and transformer-based detectors have become the dominant approach to automated surface inspection in manufacturing~\citep{tabernik2020segmentation,he2019end}.
Public benchmarks such as NEU-DET for hot-rolled steel~\citep{he2019end} and PCB-oriented defect datasets for electronic assembly~\citep{huang2019pcb} enable reproducible comparison across methods and defect types.
One-stage detectors, including the YOLO family~\citep{redmon2016you,sapkota2025ultralytics}, are widely adopted in industrial settings because they offer a favorable accuracy--speed trade-off and straightforward deployment on production lines.

Most published systems assume access to hundreds or thousands of labeled images per defect class.
In practice, new products, rare failure modes, and line changeovers often provide only a handful of annotated examples before inspection must go live.
Under such conditions, standard detectors trained on the available few labels tend to exhibit low recall on thin, low-contrast, or small defects, even when the same architecture performs well with full training data~\citep{he2019end,huang2019pcb}.
Recent industrial studies therefore emphasize data-efficient training, lightweight backbones, and improved sampling rather than solely scaling model capacity~\citep{tabernik2020segmentation}.
Nevertheless, methods that explicitly encode \emph{boundary geometry} when labeled images are scarce remain comparatively underexplored in the inspection literature.

\subsection{Few-shot and low-data object detection}

Few-shot object detection aims to generalize to novel categories or severely limited training sets~\citep{kang2019few,wu2020meta}.
Meta-learning, metric learning, and episodic training have been proposed to transfer knowledge from base classes to novel ones~\citep{yan2019meta,fan2020few}.
These approaches can be effective on natural-image benchmarks, but they often require carefully designed base/novel splits, additional pre-training stages, or architectural changes that complicate deployment on existing inspection pipelines.

An alternative line of work keeps the detector fixed and improves \emph{training data} under a standard supervised protocol: the model, loss, and inference stack remain unchanged, while the labeled set is expanded or enriched~\citep{ghiasi2021simple,shorten2019survey}.
This perspective is attractive for industrial inspection, where engineers typically want to retain an established YOLO-based workflow and only augment the offline training corpus.

\subsection{Data augmentation for object detection}

Data augmentation is central to robust visual inspection, especially when labels are expensive~\citep{shorten2019survey,perez2017effectiveness}.
In modern detection frameworks, random photometric and geometric transforms (flipping, scaling, color jitter, mosaic-style composition) are applied online during training~\citep{bochkovskiy2020yolov4,sapkota2025ultralytics}.
Copy-paste and synthetic mixing strategies insert object instances into new backgrounds and have shown strong gains on COCO-style datasets~\citep{ghiasi2021simple,ge2021yolox}.

Industrial defect images differ from natural scenes in several ways.
Defects are often defined by \emph{boundary continuity}, elongation, or local curvature on homogeneous or repetitive backgrounds, rather than by distinctive object-level appearance alone.
Generic augmentations improve diversity but do not necessarily emphasize contour geometry.
Edge-based or attention-based enhancements exist~\citep{tabernik2020segmentation}, yet they typically rely on hand-crafted edge maps or local filters rather than a principled path-level description of boundary shape.
A systematic way to derive augmentation signals from the \emph{geometry of extracted contours}, while keeping labels and detector architecture fixed, is still missing in few-shot industrial detection.

\subsection{Path signature and rough path methods}

Rough path theory represents continuous trajectories through their \emph{signature}, a sequence of iterated integrals that captures path geometry in a structured and hierarchical manner~\citep{lyons1998,lyons2007differential}.
The signature is parameterization-invariant up to tree-like retracing and satisfies factorial decay of higher-order terms, so truncated signatures provide stable finite-dimensional summaries~\citep{hambly2010uniqueness,lyons2007differential}.
In two dimensions, the antisymmetric part of the second-order signature is related to the L\'evy area and encodes oriented enclosed area along a path (Figure~\ref{fig:levy_area}), which is sensitive to local turning and winding.

Path signatures have been applied successfully in time-series modeling, finance, and sequential learning~\citep{chevyrev2016,morrill2020generalised,guo2026signature,kuang2026financial}.
Most applications treat signatures as \emph{feature vectors} appended to downstream models.
Their use as \emph{spatial augmentation signals} for image-based object detection---in particular, for constructing second training views from contour paths in few-shot industrial inspection---has received limited attention.

As mentioned above, industrial YOLO-based detectors and public benchmarks such as NEU-DET and PCB-Defect support accurate inspection when labels are abundant, yet performance drops sharply on thin or low-contrast defects once only a handful of annotated images are available.
Few-shot and meta-learning methods mitigate label scarcity but often rely on episodic training, base/novel splits, or architectural changes that are difficult to adopt on established production pipelines.
Generic detection augmentations and copy-paste-style mixing improve exposure but do not explicitly encode \emph{boundary geometry}; edge-based enhancements remain largely filter- or attention-driven rather than path-level.
Therefore, we introduce rough path signatures to few-shot industrial defect detection as label-preserving, geometry-aware augmentation signals derived from Canny-extracted edge contours, without modifying the detector architecture or the standard supervised training protocol.

\section{Theory}
\label{sec:theory}
Rough path theory provides a coordinate-free (parameterization-invariant) description of continuous trajectories through the \emph{signature} of a path~\citep{lyons1991non,lyons2002system}.
In this work, edge contours extracted from training images are treated as ordered paths in $\mathbb{R}^2$; their truncated second-order signatures supply geometry-sensitive descriptors that are later mapped to spatial augmentation signals (Section~\ref{sec:method}).

\subsection{Truncated signature}

Let $J=[0,T]$ be a compact time interval and let $X:J\to\mathbb{R}^d$ be a continuous path of bounded variation.
Write $T^n(\mathbb{R}^d)$ for the truncated tensor algebra up to degree~$n$.
The \emph{signature} of $X$ over $J$ is the graded sequence
\begin{equation}
S(X)_J
=
\Bigl(
1,\;
X_J^1,\;
X_J^2,\;
\ldots
\Bigr)
\in T(\mathbb{R}^d),
\end{equation}
where the level-$n$ term is the iterated integral
\begin{equation}
X_J^n
=
\idotsint\limits_{t_1<\cdots<t_n;\; t_i\in J}
\mathrm{d}X_{t_1}\otimes\cdots\otimes \mathrm{d}X_{t_n}
\in \left(\mathbb{R}^d\right)^{\otimes n}.
\end{equation}
The \emph{truncated signature of order $n$} is
$S_n(X)_J=(1,X_J^1,\ldots,X_J^n)\in T^n(\mathbb{R}^d)$.
In practice only finitely many terms can be stored and computed; the truncation order is therefore chosen deliberately rather than taken to infinity.

Two properties motivate the use of signatures as geometric path features.

\begin{thm}[Uniqueness up to tree-like equivalence]
\label{thm:uniqueness}
Let $X$ be a path of bounded variation.
Then $S(X)$ determines $X$ up to tree-like equivalence~\citep{hambly2010uniqueness}.
\end{thm}

\noindent
Informally, tree-like paths retrace themselves and cancel out; Theorem~\ref{thm:uniqueness} states that, modulo such retracing, the signature encodes the geometry of the underlying trajectory.
For defect contours---which are open, non-tree-like boundary curves---this makes the signature a faithful summary of boundary shape rather than a mere list of point coordinates.

\begin{thm}[Factorial decay of signature terms]
\label{thm:decay}
Let $X$ be a $d$-dimensional path of bounded variation and let
$1\le i_1,\ldots,i_n\le d$.
Then
\begin{equation}
\left\|
\idotsint\limits_{t_1<\cdots<t_n;\; t_i\in J}
\mathrm{d}X_{t_1}^{i_1}\cdots \mathrm{d}X_{t_n}^{i_n}
\right\|
\le
\frac{\Vert X\Vert_1^{\,n}}{n!},
\end{equation}
where $\Vert X\Vert_1$ denotes the total variation of $X$ on $J$~\citep{lyons2007differential}.
\end{thm}

\noindent
Theorem~\ref{thm:decay} shows that higher-order signature terms decay factorially in magnitude.
Retaining terms up to a fixed low order therefore incurs controlled information loss.
For the contour paths considered here, we use \emph{second-order} signatures only: they already capture the principal geometric content relevant to local boundary curvature and enclosed area, while avoiding the feature explosion and computational cost of higher orders.

\subsection{Second-order signature in two dimensions}

\begin{figure}[ht]
    \centering
    \includegraphics[width=0.55\textwidth]{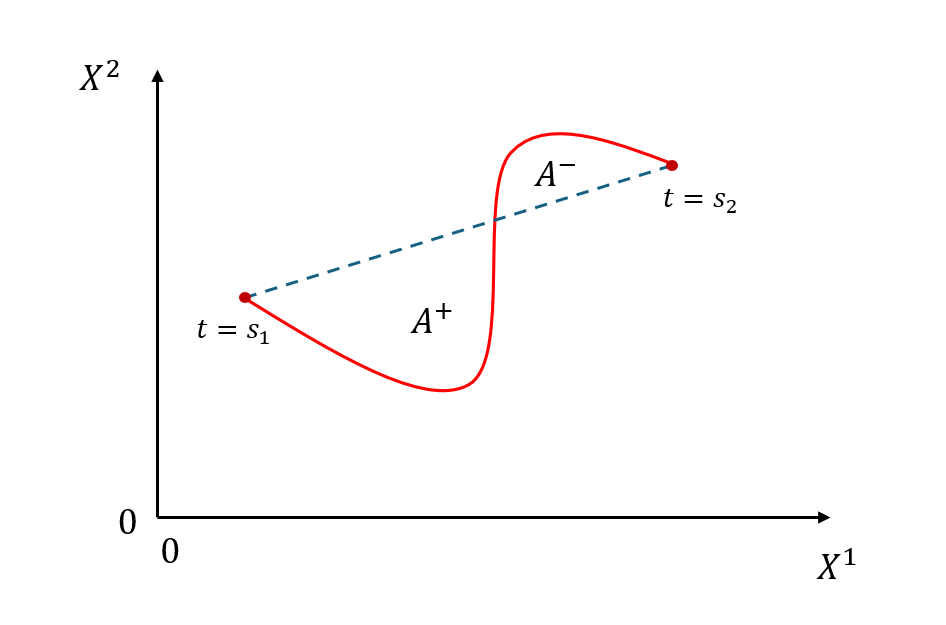}
    \caption{Geometric interpretation of the L\'evy area $A_{s,t}^{(i,j)}$ for a planar path.}
    \label{fig:levy_area}
\end{figure}

Industrial defect contours are represented as planar paths $X:J\to\mathbb{R}^2$.
Write $X_{s,t}^{(i)}$ for the increment of coordinate $i$ along $[s,t]\subset J$, and
\begin{equation}
X_{s,t}^{(i,j)}
=
\int_{s<u_1<u_2<t}
\mathrm{d}X_{u_1}^{(i)}\,\mathrm{d}X_{u_2}^{(j)},
\qquad i,j\in\{1,2\},
\end{equation}
for the second-order iterated integral.
The second-order term admits the decomposition
\begin{equation}
X_{s,t}^{(i,j)} = A_{s,t}^{(i,j)} + C_{s,t}^{(i,j)},
\label{eq:AC-decomp}
\end{equation}
where the antisymmetric part
\begin{equation}
A_{s,t}^{(i,j)}
=
\frac{1}{2}\left(
X_{s,t}^{(i,j)} - X_{s,t}^{(j,i)}
\right)
\label{eq:levy-area}
\end{equation}
is the \emph{L\'evy area} of the path: it measures oriented enclosed area between the trajectory and its chord (Figure~\ref{fig:levy_area}).
The symmetric part $C_{s,t}^{(i,j)}$ depends on path increments alone and reflects displacement along each coordinate.

\noindent
For closed or nearly closed boundary segments, $A$ is sensitive to local turning and winding of the contour, whereas first-order terms record only net displacement.
This distinction is useful for defect detection: elongated scratches, cracks, and irregular boundaries differ not only in location but also in how their boundaries bend and enclose area.
The symmetric component $C$ can be written explicitly in terms of increments; for example,
\begin{equation}
C_{s,t}
=
\begin{bmatrix}
\tfrac{1}{2}\bigl(X_{s,t}^{(1)}\bigr)^2 &
\tfrac{1}{2}X_{s,t}^{(1)}X_{s,t}^{(2)} \\[4pt]
\tfrac{1}{2}X_{s,t}^{(1)}X_{s,t}^{(2)} &
\tfrac{1}{2}\bigl(X_{s,t}^{(2)}\bigr)^2
\end{bmatrix}.
\end{equation}

In implementation, each resampled contour is centroid-centered and scale-normalized before signature evaluation.
Second-order terms are computed with the \texttt{iisignature} library~\citep{chevyrev2016}.
Following Eq.~\eqref{eq:levy-area}, the antisymmetric response associated with each contour is aggregated into a spatial map over the image domain; Section~\ref{sec:method} describes how this map drives SIG-AUG and SGAA.

\section{Method}
\label{sec:method}

We propose \emph{rough path signature-guided geometry augmentation} (RPS-GA), a data-level framework for few-shot industrial defect detection built on a fixed object detector.
For each labeled training image, RPS-GA extracts ordered contour paths, evaluates a truncated second-order path signature response map $M(\mathbf{I})$, and constructs one additional labeled training view through either SIG-AUG or SGAA.
Models are trained on paired original and augmented images; validation and testing use unmodified images only.

\begin{figure}[H]
    \centering
    \includegraphics[width=\textwidth]{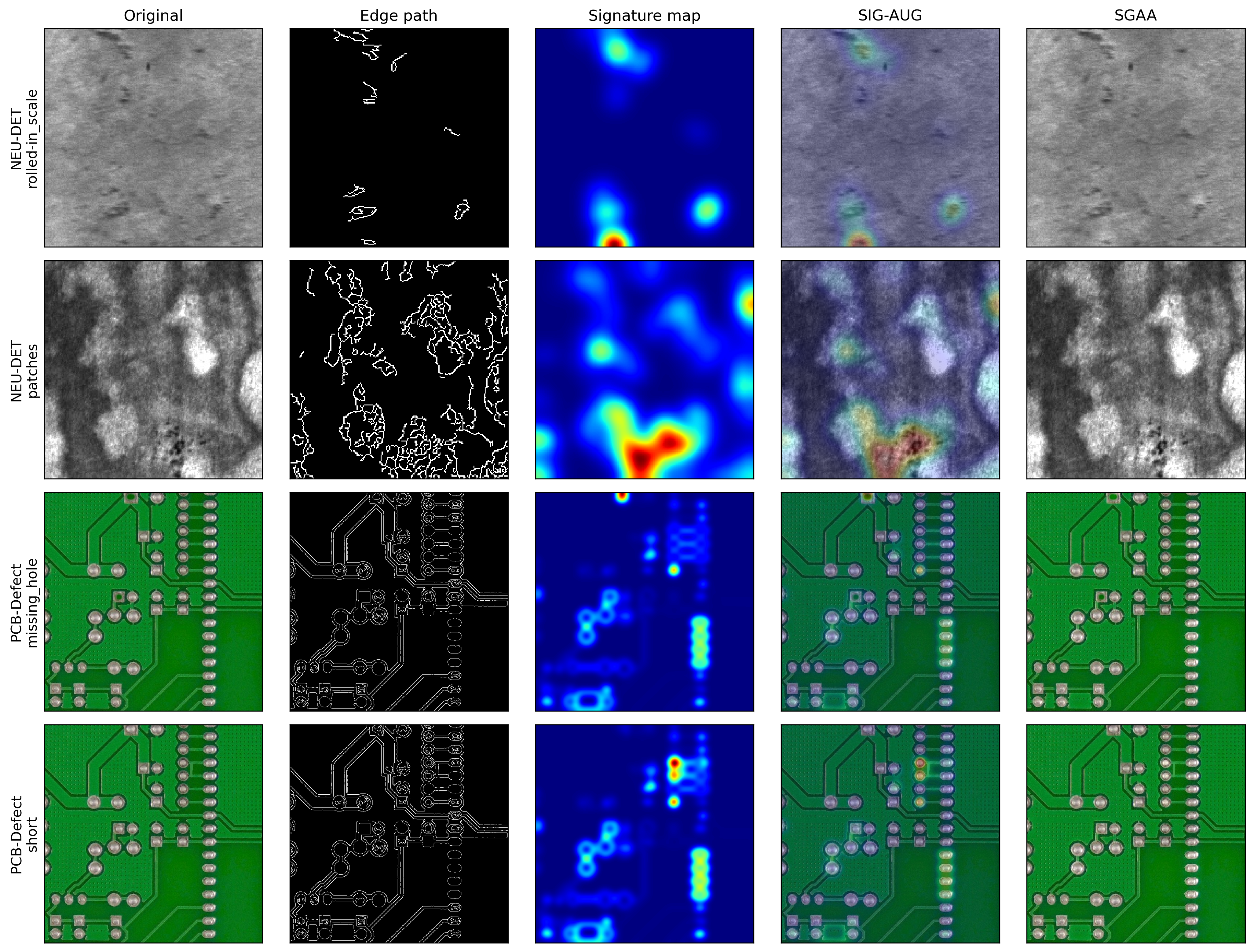}
    \caption{Illustration of the RPS-GA pipeline on NEU-DET and PCB-Defect samples.}
    \label{fig:method_visualization}
\end{figure}

Fig.~\ref{fig:method_visualization} summarizes the end-to-end workflow. Each row corresponds to one dataset (NEU-DET or PCB-Defect) and shows, from left to right:
the original training image,
the extracted edge contour used as an ordered path,
the second-order signature response map $M(\mathbf{I})$,
the SIG-AUG augmented view,
and the SGAA augmented view.
The signature map highlights boundary-dominated structures; SIG-AUG and SGAA produce two complementary second training views from the same map (Sections~\ref{sec:signature_map}).
The map aggregates second-order signature responses along Canny contours over the full image; high-response regions indicate geometrically salient boundary segments rather than a class-specific segmentation mask.

All experiments employ YOLOv8n without architectural modification; the signature theory is given in Section~\ref{sec:theory}.

\subsection{Datasets and inspection setting}
We apply RPS-GA to two public industrial defect datasets that differ in imaging modality and defect appearance.\\
\textbf{NEU-DET.}
The NEU surface defect database contains hot-rolled steel-strip images with six defect categories: crazing, inclusion, patches, pitted surface, rolled-in scale, and scratches.
Defects are often boundary-dominated and appear at low contrast on homogeneous metal backgrounds, which makes them difficult to learn from only a few labeled examples.\\
\textbf{PCB-Defect.}
The PCB defect dataset contains printed-circuit-board images with six defect types: mouse bite, spur, missing hole, short, open circuit, and spurious copper. The dataset is based on the public PCB defect benchmark PKU-Market-PCB~\citep{huang2019pcb}.
In our experiments, we use a publicly released YOLO-format split derived from this benchmark, rather than the original 1,386-image release.
Compared with NEU-DET, PCB defects are typically smaller and embedded in complex repetitive textures, posing a different few-shot generalization challenge.\\
Both datasets provide axis-aligned bounding-box annotations in YOLO format, enabling direct training and evaluation with YOLOv8n.
Together they cover metallic-surface inspection and electronic-assembly inspection, two common industrial visual-inspection settings in which labeled defect data are costly to collect.

\subsection{Signature response map construction}
\label{sec:signature_map}
The same map $M(\mathbf{I})$ is shared by SIG-AUG and SGAA.
Let $\mathbf{I}\in\mathbb{R}^{H\times W\times 3}$ denote a training image.\\
\textbf{Edge paths.}
The image is converted to grayscale, smoothed with a Gaussian filter, and processed with Canny edge detection (thresholds $50$ and $150$).
Connected edge components are traced as ordered contours.
Each contour is resampled to at most $128$ points along arc length, yielding a path $\gamma=\{(x_t,y_t)\}_{t=1}^{T}$.
Short or unstable contours are discarded.\\
\textbf{Path preprocessing.}
Before signature evaluation, each retained path is centroid-centered and scale-normalized so that geometric descriptors are comparable across defects of different sizes and image locations.\\
\textbf{Second-order signature response.}
For each normalized contour, the truncated signature of order two is computed with \texttt{iisignature}~\citep{chevyrev2016}.
Following the antisymmetric decomposition in Eq.~\eqref{eq:levy-area}, a scalar response proportional to the L\'evy-area term is assigned to the contour; in implementation this is obtained from the antisymmetric second-order signature components (equivalently, the difference between the cross-channel second-order terms).
The scalar is further weighted by contour length, painted along the contour pixels, and accumulated over all retained contours to form a raw spatial map.
The map is Gaussian-smoothed ($\sigma=9$), clipped to $[0,1]$, and min--max normalized to obtain $M(\mathbf{I})\in[0,1]^{H\times W}$.
High-response regions highlight boundary segments whose second-order path geometry is informative. This step is independent of the downstream augmentation operator.

\subsection{SIG-AUG}
SIG-AUG (\emph{signature-guided augmentation}) converts $M(\mathbf{I})$ into an explicit geometric cue by overlaying a pseudo-color heatmap on the original image:
\begin{equation}
\mathbf{I}_{\mathrm{sig}}
=
(1-\alpha)\,\mathbf{I}
+
\alpha\,\mathcal{H}\!\left(M(\mathbf{I})\right),
\label{eq:sig_aug}
\end{equation}
where $\mathcal{H}(\cdot)$ applies the JET colormap to $M(\mathbf{I})$ and $\alpha=0.22$.
This operator yields a second training view $\mathbf{I}_{\mathrm{sig}}$ in which signature-activated boundary regions are explicitly highlighted.

\subsection{SGAA}
SGAA (\emph{signature-guided attention augmentation}) applies the same map multiplicatively:
\begin{equation}
\mathbf{I}_{\mathrm{sgaa}}
=
\mathbf{I}\odot\left(\mathbf{1}+\beta\,M(\mathbf{I})\right),
\label{eq:sgaa}
\end{equation}
where $\odot$ denotes element-wise multiplication, $\mathbf{1}$ is the all-ones tensor, and $\beta=0.35$.
Compared with SIG-AUG, SGAA preserves the original color distribution while amplifying geometrically salient boundary structures.
Both operators use the identical signature pipeline in Section~\ref{sec:signature_map}; they differ only in how $M(\mathbf{I})$ is fused with $\mathbf{I}$.

\subsection{Few-shot training protocol}
For each dataset, a few-shot subset is formed by sampling $K$ labeled images per defect class ($K\in\{5,10,20,50\}$ in the main experiments).
RPS-GA doubles the training set by pairing every original image with one augmented view:
\begin{itemize}
  \item original images $\{\mathbf{I}_i\}$ with unchanged bounding-box labels;
  \item one augmented image $\{\mathbf{I}'_i\}$ per original, where $\mathbf{I}'_i$ is either $\mathbf{I}_{\mathrm{sig},i}$ or $\mathbf{I}_{\mathrm{sgaa},i}$.
\end{itemize}
Training uses paired original and augmented images, while validation and test use original images only.
Augmented images are generated offline and stored with distinct filenames; label files are copied without modification.
All compared models share the same YOLOv8n initialization, input resolution ($640\times640$), training budget (100 epochs), and dataset-specific batch size (16 for NEU-DET, 32 for PCB-Defect), with \texttt{amp=False}.
No change is made to the detector architecture or loss function.
This protocol isolates the effect of signature-guided data construction under a fixed real-time inspection backbone.

\subsection{Design rationale for few-shot inspection}
Standard detectors trained on only a few images per class often under-detect thin or low-contrast defects because the model sees insufficient boundary diversity.
RPS-GA addresses this limitation at the data level rather than by enlarging the network.
The shared map $M(\mathbf{I})$ encodes second-order boundary geometry (Section~\ref{sec:theory}), and the two operators provide complementary second views: SIG-AUG exposes geometry through pseudo-color emphasis, whereas SGAA injects geometry through multiplicative attention without altering the global appearance statistics as strongly.
Because both variants preserve the original labels and detector, they are directly deployable in existing YOLO-based inspection pipelines. Signature maps and augmented images are constructed offline, so inference at test time follows the standard YOLO pipeline without additional computation.

\section{Results}
\label{sec:results}

This section reports few-shot defect detection results on NEU-DET and PCB-Defect.
All quantitative results are evaluated on the held-out test sets.
The main comparison uses a fixed few-shot partition (single-run); stability across random partitions is assessed separately with three independent seeds at 5-, 10-, and 20-shot settings.
Unless stated otherwise, metrics are reported as mAP@0.5 and mAP@0.5:0.95 from the YOLO evaluation protocol.

\subsection{Main few-shot detection performance}
\label{sec:main_results}

Tables~\ref{tab:main_neu} and~\ref{tab:main_pcb} summarize test-set performance for YOLOv8n, SIG-AUG, and SGAA under 5-, 10-, 20-, and 50-shot training per class.

\begin{table}[H]
\centering
\caption{Few-shot detection performance on NEU-DET test set (mAP@0.5).}
\label{tab:main_neu}
\setlength{\tabcolsep}{16pt}
\begin{tabular}{l|c|c|c}
\hline
Shots per class & YOLOv8n & SIG-AUG & SGAA \\
\hline
5  & 0.114 & 0.259 & 0.275 \\
10 & 0.341 & 0.583 & 0.569 \\
20 & 0.570 & 0.654 & 0.605 \\
50 & 0.665 & 0.701 & 0.690 \\
\hline
\end{tabular}
\end{table}

\begin{table}[H]
\centering
\caption{Few-shot detection performance on PCB-Defect test set (mAP@0.5).}
\label{tab:main_pcb}
\setlength{\tabcolsep}{16pt}
\begin{tabular}{l|c|c|c}
\hline
Shots per class & YOLOv8n & SIG-AUG & SGAA \\
\hline
5  & 0.000 & 0.019 & 0.018 \\
10 & 0.086 & 0.279 & 0.299 \\
20 & 0.295 & 0.501 & 0.511 \\
50 & 0.665 & 0.680 & 0.677 \\
\hline
\end{tabular}
\end{table}

\textbf{NEU-DET.}
On the steel-surface benchmark, both proposed operators substantially outperform the YOLOv8n baseline across all shot settings.
At 5-shot, mAP@0.5 increases from 0.114 to 0.259 (SIG-AUG) and 0.275 (SGAA), indicating that geometry-aware augmentation is effective under extreme label scarcity.
The largest relative gain occurs at 10-shot, where SIG-AUG reaches 0.583 and SGAA reaches 0.569, compared with 0.341 for the baseline.
At 20-shot, SIG-AUG remains strongest (0.654), while SGAA achieves 0.605; both methods stay clearly above the baseline (0.570).
As the shot count increases to 50, all methods improve and the margin narrows (baseline 0.665; SIG-AUG 0.701; SGAA 0.690), which is expected because additional original supervision reduces the relative benefit of augmentation.

\textbf{PCB-Defect.}
On the PCB benchmark, the baseline essentially fails at 5-shot (mAP@0.5 = 0.000), whereas both SIG-AUG and SGAA enable non-trivial detection (0.019 and 0.018, respectively).
At 10-shot, SGAA achieves the best result (0.299), followed by SIG-AUG (0.279) and the baseline (0.086).
Similar trends hold at 20-shot, where SGAA (0.511) and SIG-AUG (0.501) remain well above the baseline (0.295).
At 50-shot, all methods converge to similar performance (0.665--0.680), suggesting that augmentation is most valuable in the low- and medium-shot regimes relevant to practical few-shot deployment.

\begin{figure}[H]
    \centering
    \includegraphics[width=\textwidth]{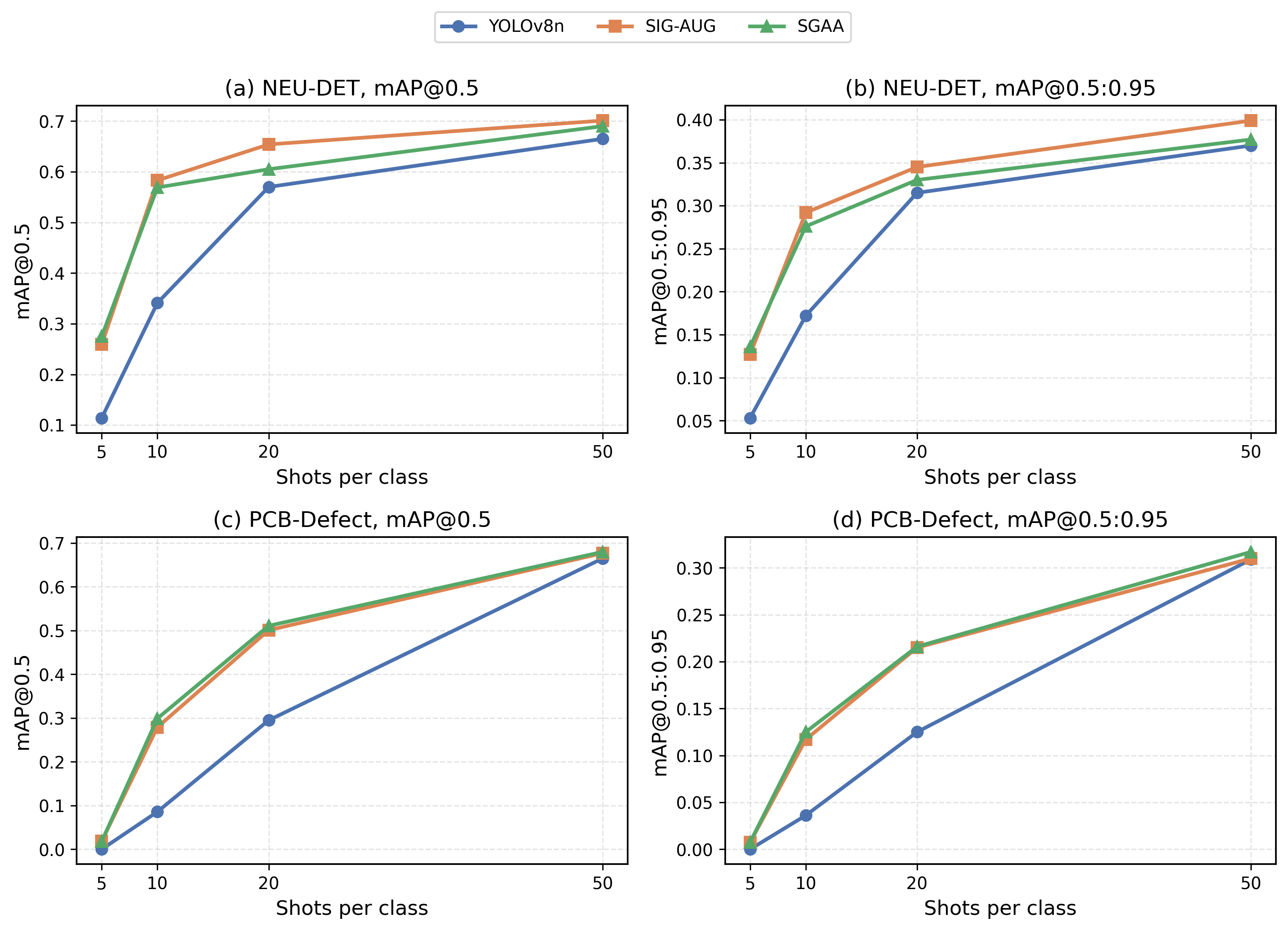}
    \caption{Few-shot detection performance on NEU-DET and PCB-Defect test sets (single-run).
    (a)--(b) NEU-DET mAP@0.5 and mAP@0.5:0.95;
    (c)--(d) PCB-Defect mAP@0.5 and mAP@0.5:0.95.}
    \label{fig:main_curves}
\end{figure}

\begin{figure}[H]
    \centering
    \includegraphics[width=\textwidth]{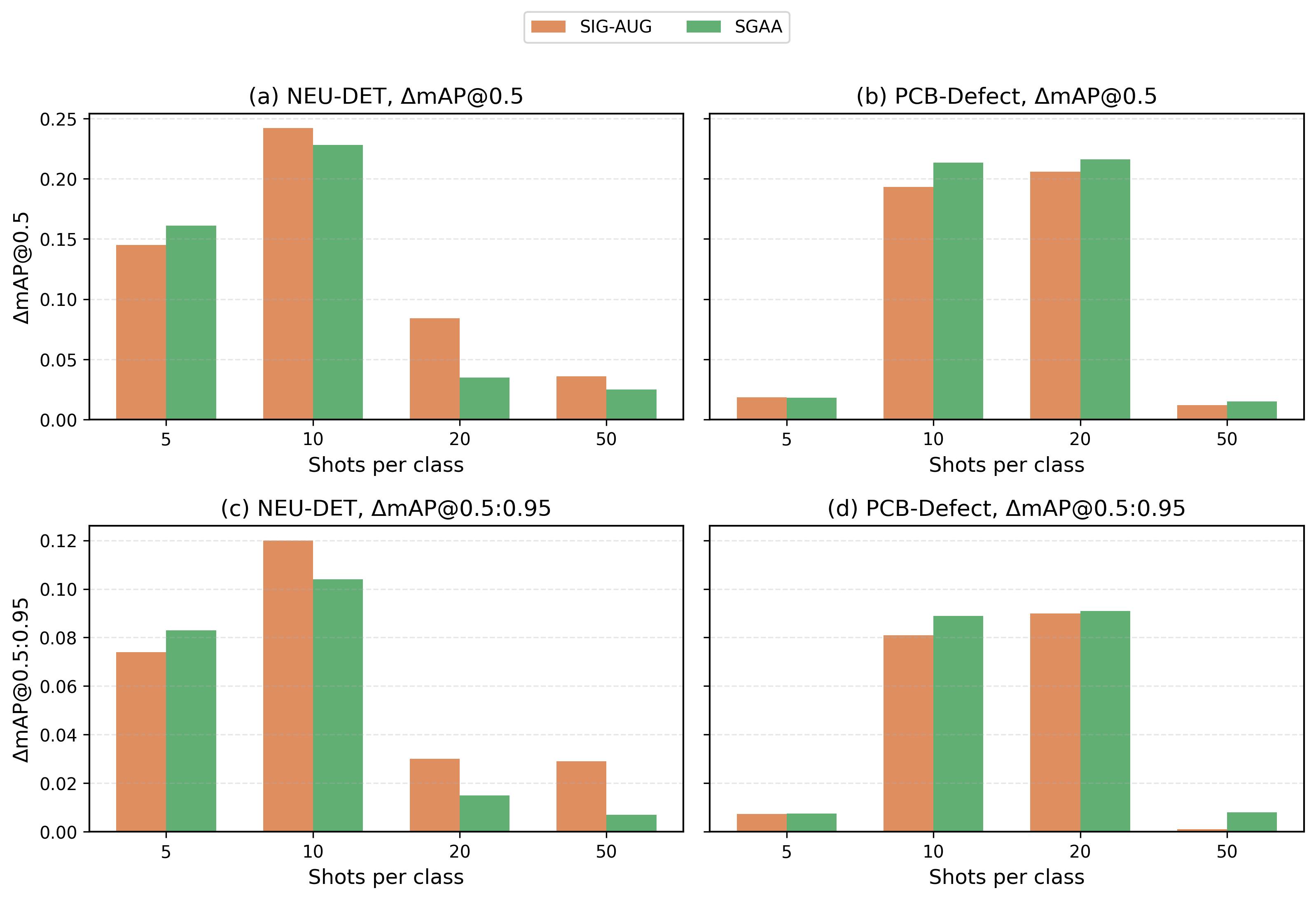}
    \caption{Improvement of SIG-AUG and SGAA over YOLOv8n baseline on NEU-DET and PCB-Defect (single-run, mAP@0.5).}
    \label{fig:improvement_bars}
\end{figure}

Figure~\ref{fig:main_curves} shows the corresponding mAP curves, and Figure~\ref{fig:improvement_bars} reports the absolute improvement over the baseline.

Fig.~\ref{fig:main_curves}(a)--(b) and (c)--(d) confirm these patterns for both mAP@0.5 and the stricter mAP@0.5:0.95 metric.
Fig.~\ref{fig:improvement_bars} shows that the largest $\Delta$mAP gains occur at 5- and 10-shot on NEU-DET and at 10- and 20-shot on PCB-Defect, while improvements become modest at 50-shot.

\subsection{Stability across random few-shot partitions}
\label{sec:stability}

\begin{figure}[H]
    \centering
    \includegraphics[width=\textwidth]{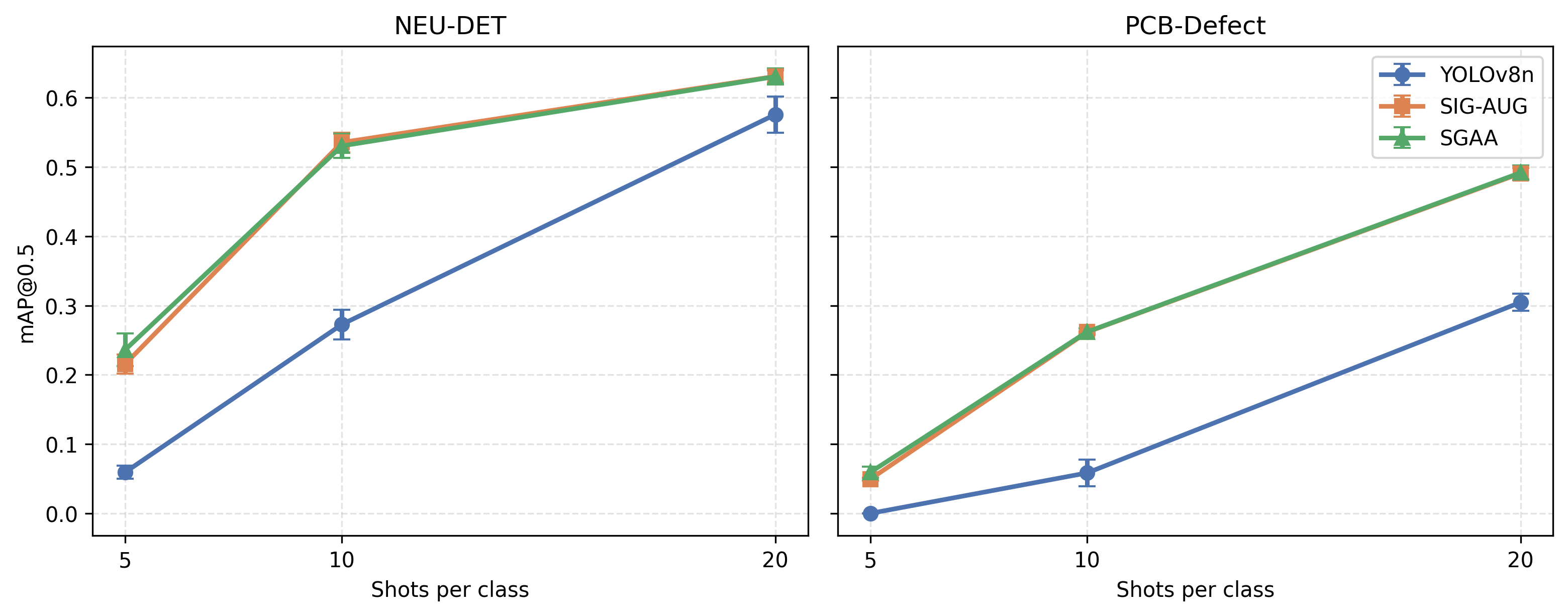}
    \caption{Multi-seed stability analysis on NEU-DET and PCB-Defect test sets (mean $\pm$ std over 3 random seeds, mAP@0.5).}
    \label{fig:stability_map50}
\end{figure}

\begin{figure}[H]
    \centering
    \includegraphics[width=\textwidth]{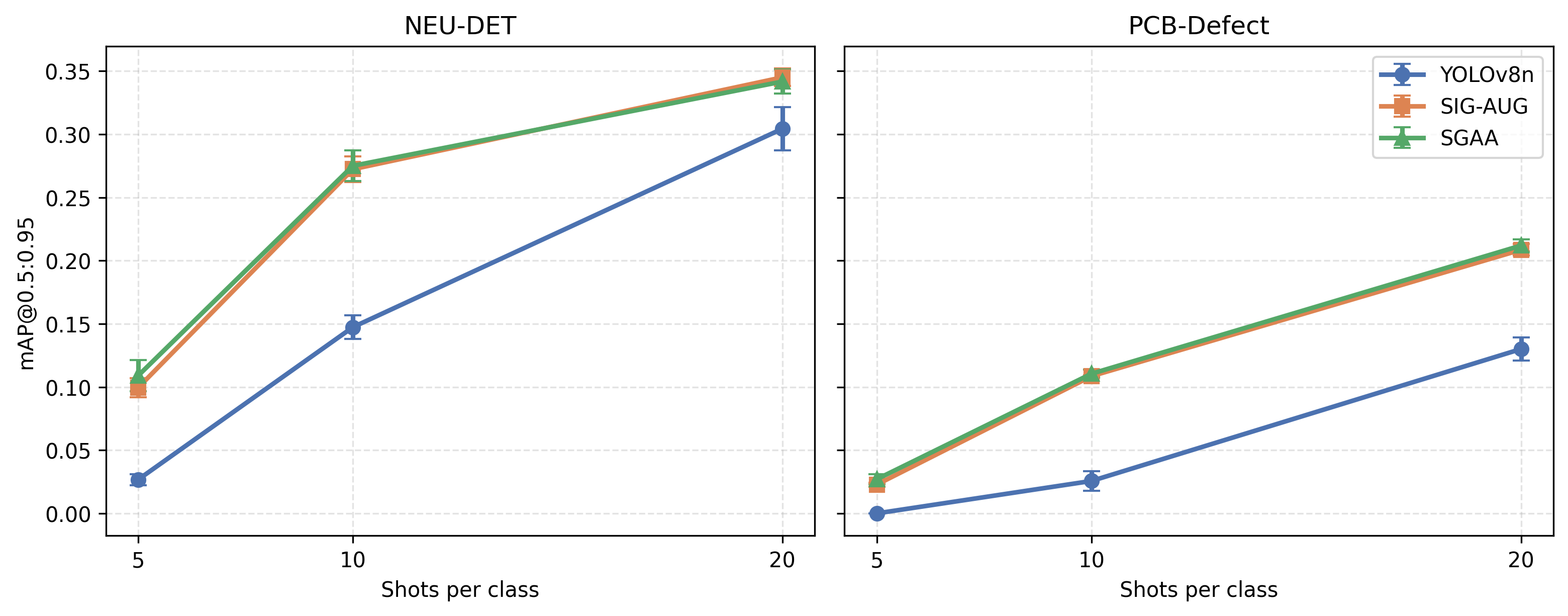}
    \caption{Multi-seed stability analysis on NEU-DET and PCB-Defect test sets (mean $\pm$ std over 3 random seeds, mAP@0.5:0.95).}
    \label{fig:stability_map5095}
\end{figure}

To verify that the main gains are not caused by a particular few-shot sampling, we repeat the experiments with three independent random partitions at 5-, 10-, and 20-shot settings.
Tables~\ref{tab:stability_neu} and~\ref{tab:stability_pcb} report mean~$\pm$~standard deviation of mAP@0.5 over the three seeds; Figure~\ref{fig:stability_map50} and~\ref{fig:stability_map5095} show the same trends for mAP@0.5 and mAP@0.5:0.95.

\begin{table}[H]
\centering
\caption{Multi-seed stability on NEU-DET test set (mean $\pm$ std over 3 seeds, mAP@0.5).}
\label{tab:stability_neu}
\setlength{\tabcolsep}{16pt}
\begin{tabular}{l|c|c|c}
\hline
Shots per class & YOLOv8n & SIG-AUG & SGAA \\
\hline
5  & $0.059 \pm 0.010$ & $0.215 \pm 0.014$ & $0.236 \pm 0.024$ \\
10 & $0.273 \pm 0.021$ & $0.535 \pm 0.014$ & $0.530 \pm 0.018$ \\
20 & $0.576 \pm 0.026$ & $0.631 \pm 0.009$ & $0.630 \pm 0.012$ \\
\hline
\end{tabular}
\end{table}

\begin{table}[H]
\centering
\caption{Multi-seed stability on PCB-Defect test set (mean $\pm$ std over 3 seeds, mAP@0.5).}
\label{tab:stability_pcb}
\setlength{\tabcolsep}{16pt}
\begin{tabular}{l|c|c|c}
\hline
Shots per class & YOLOv8n & SIG-AUG & SGAA \\
\hline
5  & $0.000 \pm 0.000$ & $0.049 \pm 0.002$ & $0.059 \pm 0.008$ \\
10 & $0.058 \pm 0.019$ & $0.262 \pm 0.001$ & $0.262 \pm 0.005$ \\
20 & $0.305 \pm 0.012$ & $0.491 \pm 0.008$ & $0.492 \pm 0.010$ \\
\hline
\end{tabular}
\end{table}

On NEU-DET, SIG-AUG and SGAA consistently exceed the baseline.
For example, at 10-shot the baseline mean mAP@0.5 is $0.273 \pm 0.021$, whereas SIG-AUG and SGAA reach $0.535 \pm 0.014$ and $0.530 \pm 0.018$, respectively.
At 20-shot, both augmented methods remain near 0.63 mAP@0.5 while the baseline stays at $0.576 \pm 0.026$.
The standard deviations of the augmented methods are generally small, indicating stable behavior across partitions.

On PCB-Defect, the same conclusion holds for 10- and 20-shot settings.
At 10-shot, SIG-AUG and SGAA both achieve mean mAP@0.5 of approximately 0.262, compared with $0.058 \pm 0.019$ for the baseline.
At 20-shot, the augmented methods reach about 0.49--0.49 mAP@0.5, while the baseline remains at $0.305 \pm 0.012$.
At 5-shot, absolute mAP values are low for all methods because only 30 training images are available in total; nevertheless, RPS-GA still enables detectable performance where the baseline mean is zero.

Overall, the multi-seed analysis supports that the proposed geometry-guided augmentation improves few-shot detection reliably, not only on the fixed partition used for Tables~\ref{tab:main_neu} and~\ref{tab:main_pcb}.

\subsection{Qualitative detection results}
\label{sec:qualitative}

\begin{figure}[H]
    \centering
    \includegraphics[width=\textwidth]{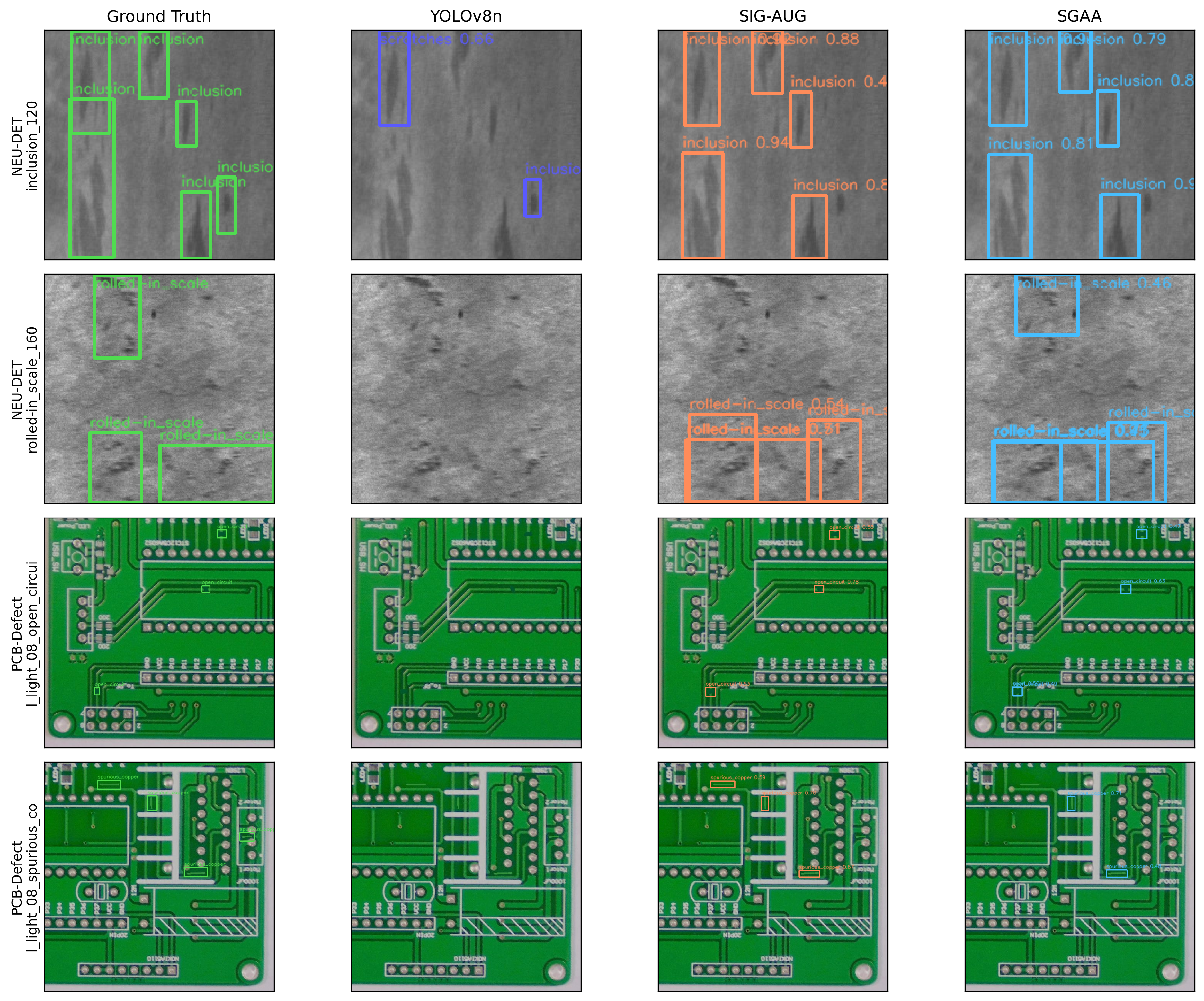}
    \caption{Qualitative detection comparison on representative test images from NEU-DET and PCB-Defect.
    Columns from left to right: ground truth, YOLOv8n baseline, SIG-AUG, and SGAA.}
    \label{fig:detection_comparison}
\end{figure}

Figure~\ref{fig:detection_comparison} compares ground-truth annotations with predictions from YOLOv8n, SIG-AUG, and SGAA on representative NEU-DET and PCB-Defect test images.

On NEU-DET, the baseline misses multiple inclusion defects and fails entirely on rolled-in scale samples, while both SIG-AUG and SGAA recover most ground-truth instances.
On PCB-Defect, the baseline often produces no detections on small open-circuit and spurious-copper defects, whereas the proposed methods localize these instances with reasonable confidence.

\begin{figure}[H]
    \centering
    \includegraphics[width=\textwidth]{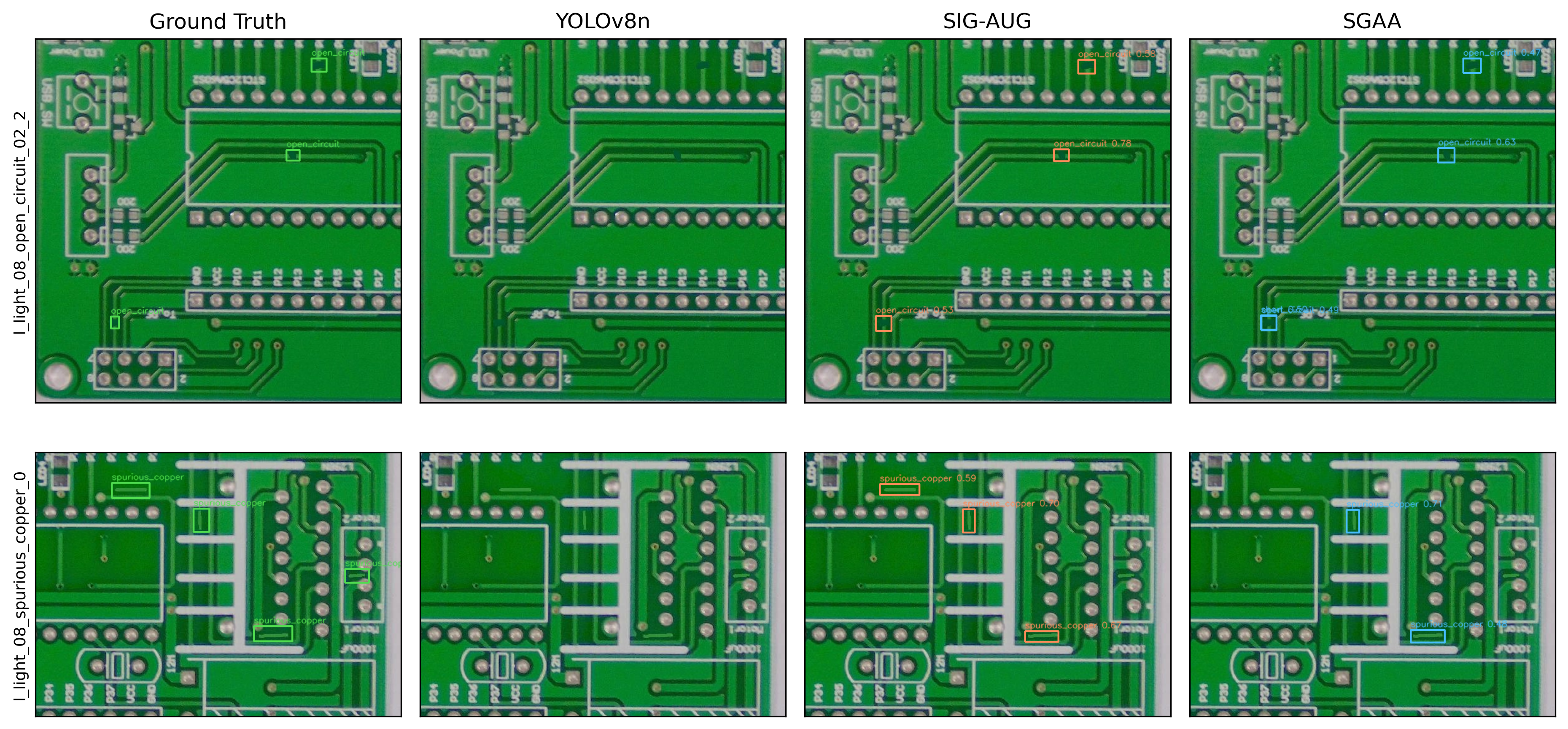}
    \caption{Zoomed qualitative comparison on PCB-Defect test images with small open-circuit and spurious-copper defects.}
    \label{fig:pcb_zoom_comparison}
\end{figure}

Figure~\ref{fig:pcb_zoom_comparison} provides a zoomed view on PCB test cases with very small defects.
These examples illustrate that signature-guided augmentation helps the detector learn boundary-related cues that are difficult to acquire from only a few original PCB images.

The qualitative examples are consistent with the quantitative trends in Tables~\ref{tab:main_neu} and~\ref{tab:main_pcb}, especially under 10- and 20-shot training.


\section{Conclusion}
\label{sec:conclusion}

This paper addressed few-shot industrial surface defect detection when only a few labeled images per class are available and an established YOLO-based inspection workflow must remain unchanged.
We proposed rough path signature-guided geometry augmentation (RPS-GA), which builds a second-order signature response map from Canny edge contours and applies SIG-AUG and SGAA to emphasize boundary-related geometry during training, while YOLOv8n and test-time inputs remain unmodified.

Experiments on NEU-DET and PCB-Defect showed that RPS-GA improves few-shot detection on both benchmarks without modifying the detector architecture.
SIG-AUG performed particularly strongly on NEU-DET at 10- and 20-shot, whereas SGAA was slightly stronger on PCB-Defect at the same settings.
Gains were largest when labeled data were scarce and became smaller as the shot count increased to 50.
Multi-seed evaluation further confirmed that these improvements are stable across different few-shot partitions.
Qualitative results also indicated that signature-guided augmentation helps recover thin, low-contrast, and small boundary-dominated defects that the baseline often misses.

Several limitations remain.
RPS-GA relies on Canny edge extraction and contour resampling, so its effectiveness may degrade when defect boundaries are weak, fragmented, or heavily obscured by background clutter.
In addition, the largest improvements occur when the number of labeled images per class is small, and the relative benefit of geometry-guided augmentation diminishes as more supervision becomes available.
Future work may examine additional industrial benchmarks, stronger boundary cues, and more adaptive signature-map construction.

\bibliographystyle{apacite}
\bibliography{bibliography.bib}

\end{document}